\documentclass[sigplan]{acmart}

\usepackage{hyperref}
\hypersetup{
    colorlinks=true,
    linkcolor=blue,
    filecolor=magenta,      
    urlcolor=cyan,
}

\setcopyright{rightsretained}

\acmConference[]{}
\copyrightyear{2018}
\acmYear{}

\begin{document}
\title{Semantic White Balance: Semantic Color Constancy Using Convolutional Neural Network}

\author{Mahmoud Afifi}

\affiliation{%
  \institution{EECS, Lassonde School of Engineering, York University, Canada}
  \postcode{M3J 1P3}
}
\email{mafifi@eecs.yorku.ca}

\begin{abstract}
The goal of computational color constancy is to preserve the perceptive colors of objects under different lighting conditions by removing the effect of color casts caused by the scene's illumination. With the rapid development of deep learning based techniques, significant progress has been made in image semantic segmentation. In this work, we exploit the semantic information together with the color and spatial information of the input image in order to remove color casts. We train a convolutional neural network (CNN) model that learns to estimate the illuminant color and gamma correction parameters based on the semantic information of the given image. Experimental results show that feeding the CNN with the semantic information leads to a significant improvement in the results by reducing the error by more than $40\%$. 
\end{abstract}


\begin{teaserfigure}
  \includegraphics[width=\textwidth]{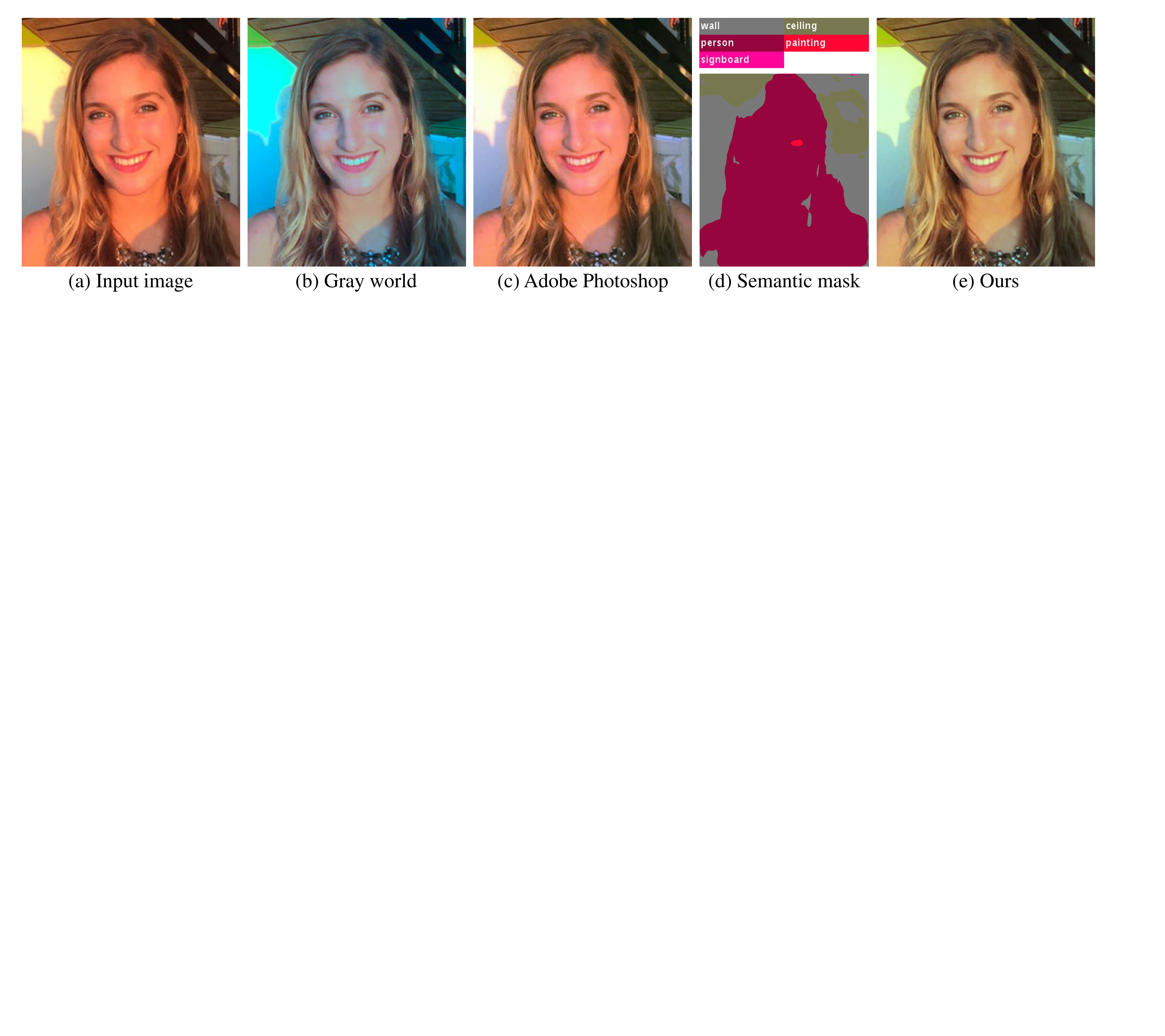}
  \caption[Color constancy using the proposed method. In (a), an input image in sRGB color space is corrected using: (b) Adobe Photoshop (auto-color correction) and (b) the proposed semantic-based color constancy method.]{Color constancy using the proposed method. In (a), an input image in sRGB color space is corrected using: (b) Adobe Photoshop (auto-color correction) and (b) the proposed semantic-based color constancy method.}
  \label{fig:teaser}
\end{teaserfigure}

\maketitle

\section{Introduction}
The goal of the computational color constancy is adjusting image colors to preserve the perceptive colors of objects under different lighting conditions by removing the illuminant color casts from the image. Typically, most algorithms are targeted towards achieving color constancy by correcting scene's illumination to be ideal white light. In the literature, these illuminant casts are usually assumed to be global and uniform (i.e., a single illuminant in the scene). Other methods went beyond this assumption by dealing with more than one light source \cite{beigpour2014multi, cheng2016two, abdelhamed2016two}. The majority of existing works focus on estimating a set of parameters to represent the scene illuminant (or illuminants in the case of more than one). Then, the correction process is assumed to be carried out by undoing the effect of that illuminant. To be able to undo the effect of the color casts, the input image should be in RAW format (i.e, both illumination estimation and correction should be applied on the linear RGB values). 
\\
In fact, the problem of color constancy is mainly related to the color distribution of the image, as shown in the literature \cite{nus, barron2015convolutional, barron2017fast, zhang2018fully}. However, the spatial information could help in improving the results. For example, objects that have innate colors could be used as guidance objects to ameliorate the illuminant estimation results. That way, Hu \textit{et al.} \cite{hu2017fc} proposed a fully convolutional color constancy method that relies on the confidence level of each patch in the image to estimate the illuminant color. The spatial information could help more, if we have some semantic information about the image content \textemdash the sky should be blue, otherwise, we can guess the level of distortion occurred in the image colors. In this work, we study the effect of feeding a convolutional neural network a semantic mask along with the input image. This mask possesses semantic information per pixel in order to enrich the CNN with useful information regarding the image content. We show that the semantic information increases the accuracy of the illuminant color estimation process, and consequently, can effectively improve the color constancy results, see Fig. \ref{fig:teaser}. 
\\
The rest of this paper is organized as follows: the proposed method is presented in Section 2 followed by the experimental results in Section 3. Eventually, the paper is concluded in Section 4. 
\begin{figure*}
  \includegraphics[width=\textwidth]{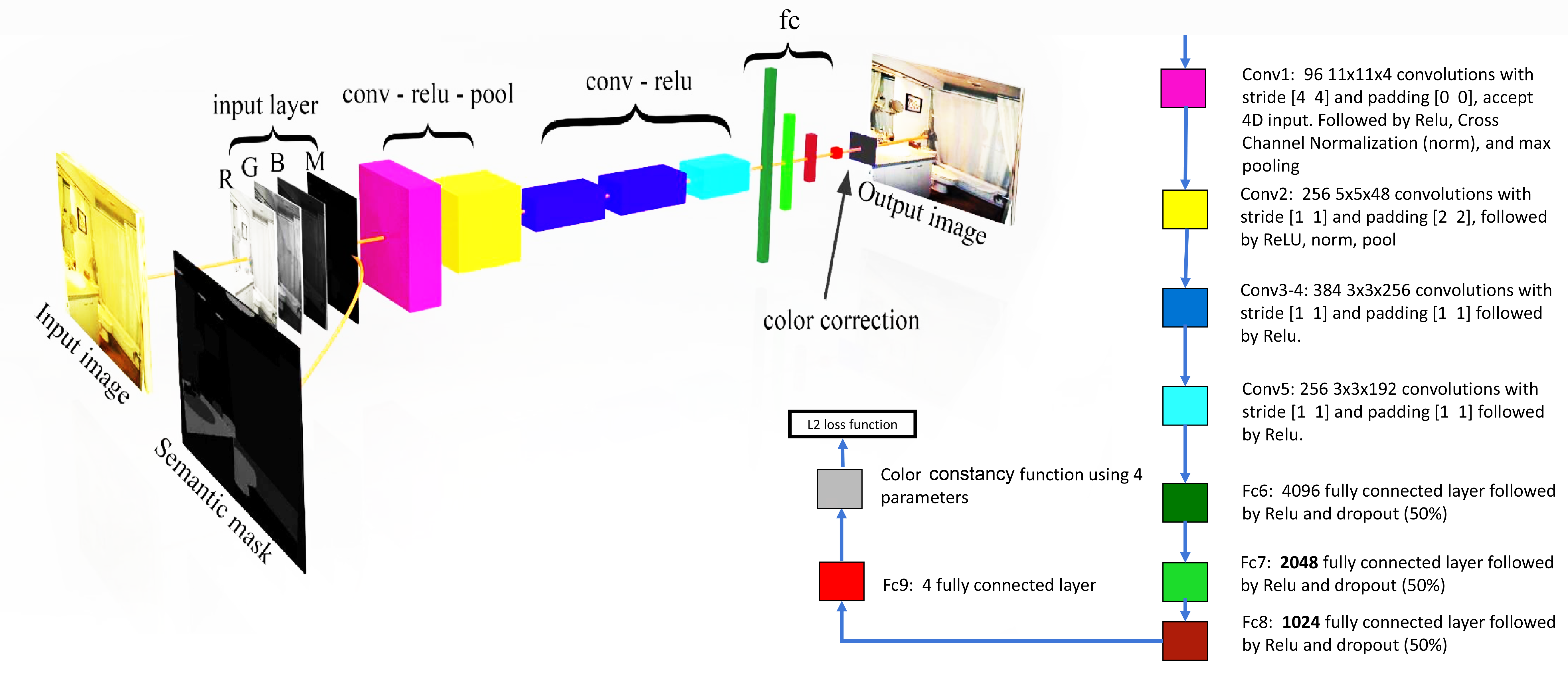}
  \caption{The proposed model receives a 4D volume as an input. The first three dimensions of the 4D volume represent the red, green, and blue color channels of the input image, while the last dimension contains the semantic mask of it.}
  \label{model}
\end{figure*}

\begin{figure*}
  \includegraphics[width=\textwidth]{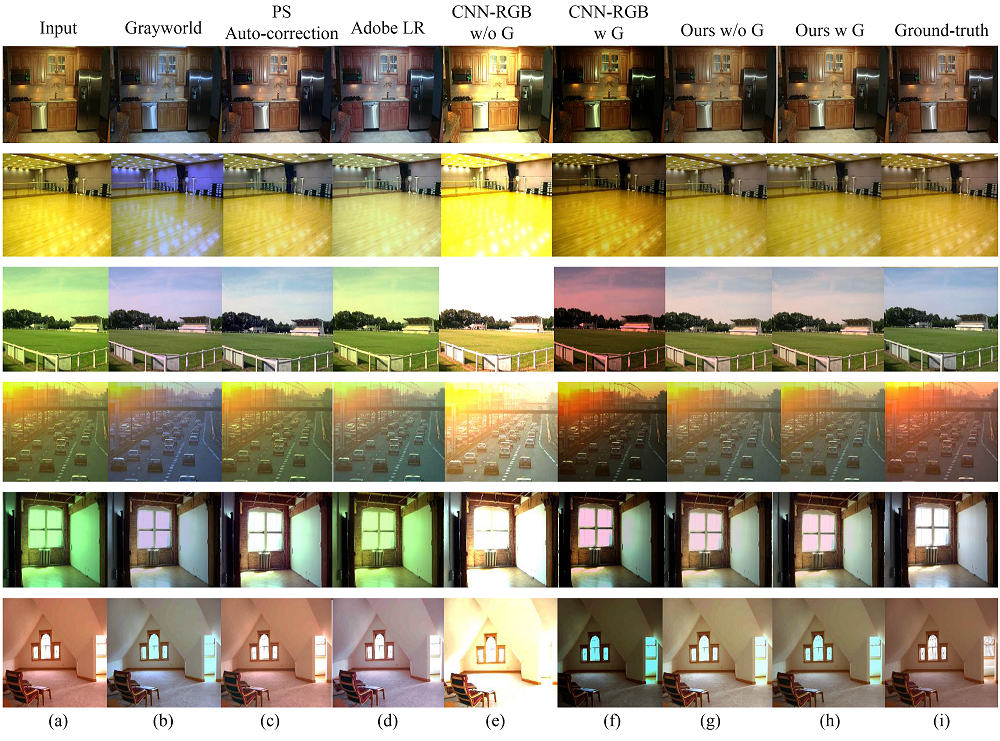}
  \caption{Results obtained using the ADE20K dataset \cite{ade20k}. (a) Input images. (b) Grey-world results \cite{buchsbaum1980spatial}. (c) Adobe auto-correction results. (d) Adobe Lightroom white balance results. (e) Results obtained using AlexNet \cite{alex} with RGB input images without using gamma correction. (f) Results of AlexNet-RGB with gamma correction. Our results with and without gamma correction are shown in (g) and (h), respectively. Finally, the ground-truth images are shown in (i). }
  \label{fig2}
\end{figure*}

\begin{figure*}
  \includegraphics[width=\textwidth]{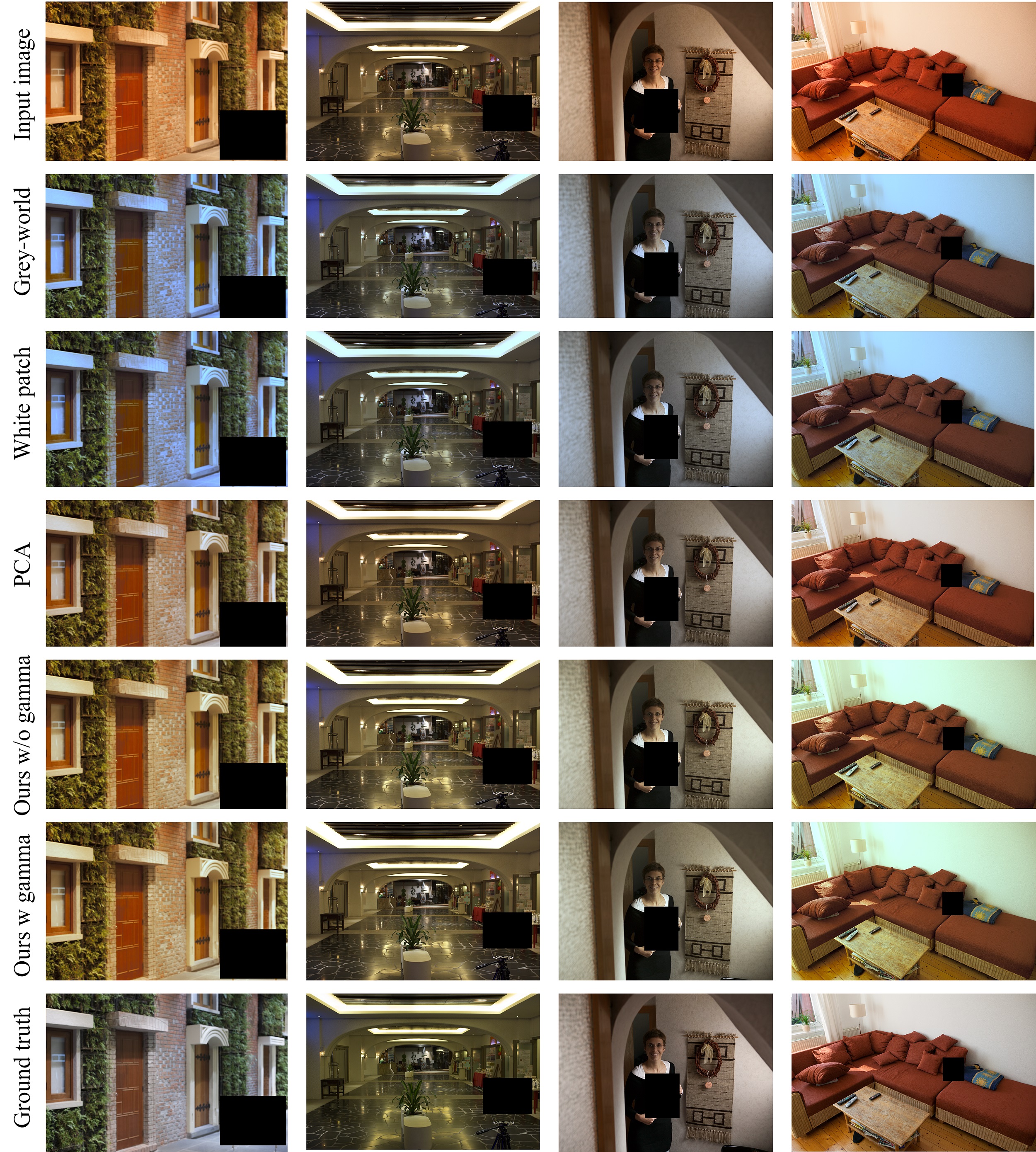}
  \caption{Comparisons with other white balance techniques. All images are in sRGB color space. Input images, from NUS and Gehler datasets \cite{nus,gehler2008bayesian}, in the first raw have been corrected using the grey-world algorithm \cite{buchsbaum1980spatial} (second row), the white patch algorithm \cite{brainard1986analysis} (third row), the PCA-based illuminant estimation \cite{nus} (fourth row), our method without gamma correction (fifth row), and our method with gamma correction (sixth row). The ground truth images are shown in the last row. }
  \label{fig2_2}
\end{figure*}

\begin{figure}
  \includegraphics[width=0.47\textwidth]{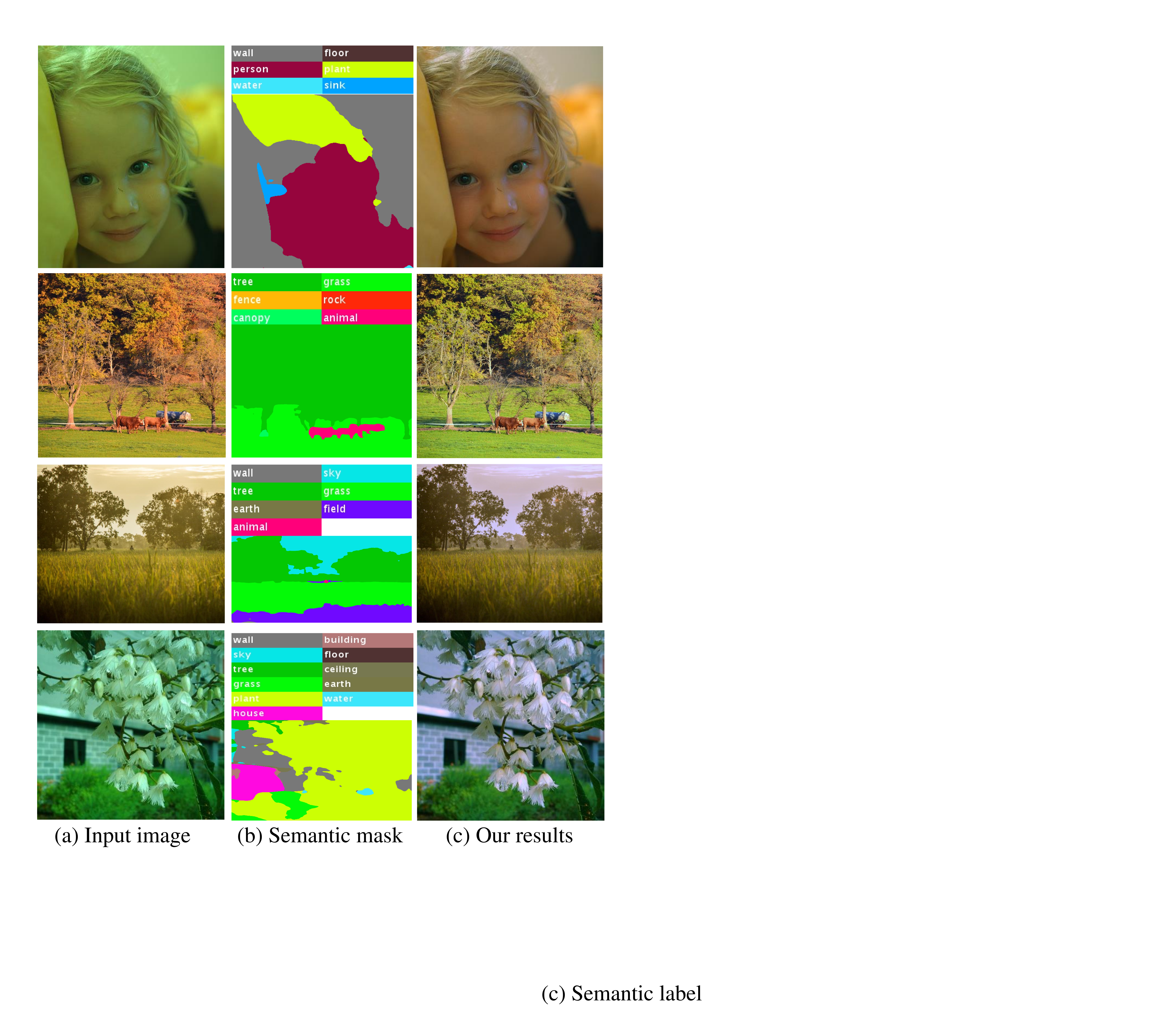}
  \caption{Quantitative results of our semantic-based white balance. (A) Input images. (B) Semantic masks obtained using Refinenet \cite{refnet}. (C) Corrected images using the proposed CNN.}
  \label{figxyz}
\end{figure}

\begin{figure}
  \includegraphics[width=0.5\textwidth]{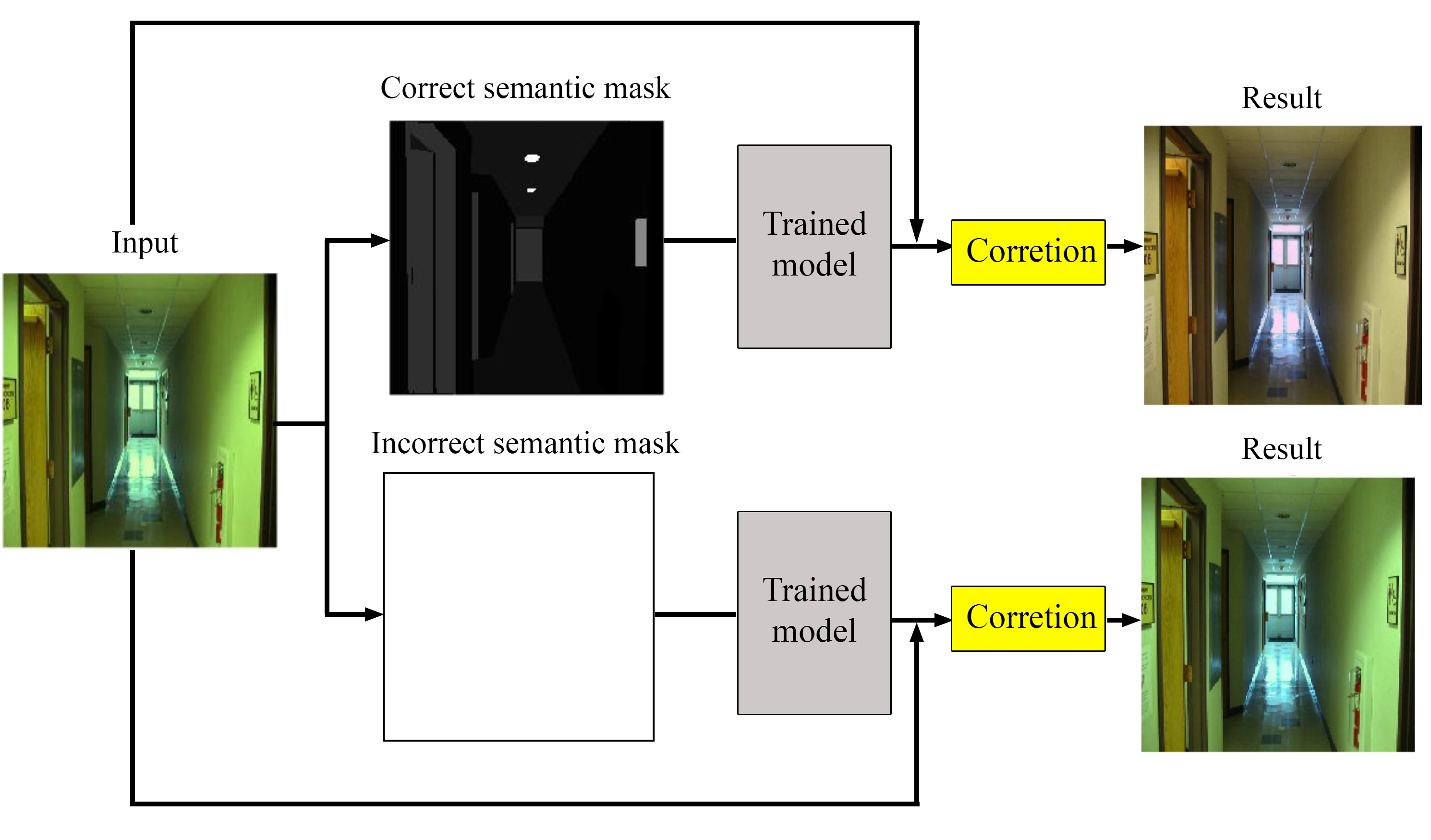}
  \caption{The effect of the semantic information on the color constancy results. We fed the trained model with two different semantic masks: a correct semantic mask (top) and an incorrect semantic mask (bottom).}
  \label{fig3}
\end{figure}

\section{Semantic Color Constancy}
 In our approach, instead of using the conventional representation of three channels, RGB, we represent images as a 4D volume, such that the first three dimensions represent the red, green, and blue color channels, and the last dimension contains the semantic mask \textemdash \space we assume that the semantic mask is given. We start with the AlexNet architecture \cite{alex}, however, we adapt it to be able to receive the 4D volume. The details of the modifications that have been applied to the AlexNet architecture are shown in Fig. \ref{model}. As illustrated in Fig. \ref{model}, we replaced the first convolutional layer (\textit{conv1}) with a new one with a filter bank consists of 96 4D filters. Furthermore, we replace the fully connected layers (\textit{fc})s with new four \textit{fc} layers, namely \textit{fc6}, \textit{fc7}, \textit{fc8}, and \textit{fc9}. Lastly, we append a regression layer to predict four parameters which are: $r$, $g$, $b$, and $\gamma$ to correct the given image. The correction is performed using the following equation: 
\begin{equation}
\hat{\mathbf{I}} = (\mathbf{I}\hat{\mathbf{M}})^{1/\hat{\gamma}},
\end{equation} 
where $\hat{\mathbf{I}}$ and $\mathbf{I}$ are the corrected and input images represented as $(N\times 3)$ matrices, respectively, $\hat{\gamma}$ is the estimated gamma correction parameter, and $\hat{\mathbf{M}}$ is a diagonal matrix given by
\begin{equation}
\hat{\mathbf{M}} = \begin{bmatrix}
1/\hat{r} & 0 & 0\\ 
0 & 1/\hat{g} & 0\\ 
0 & 0 & 1/\hat{b}
\end{bmatrix},
\end{equation}
where $\hat{r}$, $\hat{g}$, and $\hat{b}$ are the red, green, and blue values of the estimated illuminant color, respectively.
\\
\textbf{Training}
We train the CNN using the mini-batch stochastic gradient descent with momentum ($\beta=0.95$) and learning rate $\lambda = 10^{-5}$ for 500 epochs. The value of $\lambda$ is decreased each 10 epochs by $0.1$. For the new layers, we use a higher learning rate $\lambda_2$, where $\lambda_2 = \lambda \times 50$. From \textit{conv2} to \textit{conv5}, the weights are initialized with the pre-trained AlexNet's weights. The weights of the first convolutional layer of AlexNet are used to initialize the first 3 dimensions of each filter in \textit{conv1}. For the last dimension of \textit{conv1}, however, we use average filter (i.e., initialized with 1/11) to feed the mask without introducing random changes in its values, at least for the first iteration. From \textit{fc6} to \textit{fc9}, we start with random Gaussian weights. As a preprocessing step, a pixel-wise normalization is applied \textit{only} to the RGB channels of all training images.

\begin{table}[]
\centering
\caption{The average root mean square error for the testing set of the ADE20K dataset using AlexNet with and without semantic information. The first row shows the results of using images that have no gamma correction required ($\gamma = 1$). While the second row shows the results obtained using the entire testing set.}
\scalebox{1}{
\label{table1}
\begin{tabular}{c|c|c|}
\cline{2-3}
 & \textbf{w/o semantic info} & \textbf{w semantic info} \\ \hline
\multicolumn{1}{|c|}{\textbf{$\gamma=1$}} & 51.8160 & 20.1450 \\ \hline
\multicolumn{1}{|c|}{\textbf{All}} & 53.8815 & 31.1355 \\ \hline
\end{tabular}
}
\end{table}

\section{Experimental Results}
The experiments have been carried out on an Intel\textsuperscript{\textregistered} core\textsuperscript{TM} i-7 6700 @ 3.40GHz machine with 16 GB RAM and NVIDIA\textsuperscript{\textregistered} Quadro K620 graphics card. We have used the ADE20K dataset \cite{ade20k} that contains accurate semantic masks for 20,210 training images and 2,000 validation images. We have assumed that all images in the ADE20K dataset are correctly white balanced. It is worth noting that all images are in sRGB space which is an inappropriate space to apply white balance; however, we used it because of the absence of RAW image datasets provided with semantic masks. In other words, we assume that there is no camera color style applied to get the sRGB images, and the gamma correction is performed using the Adobe RGB (1998) standard \cite{headquarters2005adobe}. The Matlab source code is available on-line \footnote{Matlab code: \hyperlink{https://goo.gl/sdqCyv}{https://goo.gl/sdqCyv}} to encourage reproducibility of results. 
\\
\textbf{Data Augmentation}
As aforementioned, we assume that the sRGB images in the ADE20K dataset are corrected white balance. Consequently, for each image in both training and testing sets, we have generated 769 synthetic images by applying: 1) wrong white balance correction (randomly applied, such that $r$, $g$, and $b$ $\in$ $[0.7, 1.3]$), 2) wrong gamma correction ($\gamma \in [0.85,1.15]$), and 3) spatial data augmentation. Although the problem is related to the image colors, spatial data augmentation was applied to enrich the learning process, as we use the same image with different white balance corrections. For each image, the 769 synthesized images $\{\mathbf{I}_i^{'}\}_{i=1}^{769}$ have been generated as shown in the following equation
\begin{equation}
\mathbf{I}_i^{'} = (\mathbf{I}\mathbf{M}_i)^{\gamma_i},
\end{equation}

\begin{equation}
\mathbf{M}_i = \begin{bmatrix}
r_i & 0 & 0\\ 
0 & g_i & 0\\ 
0 & 0 & b_i
\end{bmatrix}.
\end{equation}
\\
\textbf{Quantitative Results:} We compared the proposed method against the regular AlexNet without using the semantic information. We have used the same training setup to get a fair comparison. Table \ref{table1} shows the average root mean square error (RMSE) per pixel for the testing set with and without the semantic information. The RMSE between the corrected image and the ground truth image is given by
\begin{equation}
RMSE = \sqrt{\frac{\sum_{i=1}^{N}(\hat{\mathbf{I}}- \mathbf{I}))^2}{N}},
\end{equation} 
where $N$ is the number of pixels in the image. As shown in Table \ref{table1}, the semantic information improves the results by reducing the average RMSE by more than $40\%$. 
\\
\textbf{Qualitative Results:} As shown in Fig. \ref{fig2}, the proposed method achieves good results compared to grey-world \cite{buchsbaum1980spatial}, Adobe Photoshop auto-color correction, Adobe Lightroom correction, and AlexNet without the semantic information. Fig. \ref{fig2_2} shows results obtained using the NUS and Gehler datasets \cite{nus,gehler2008bayesian}. The images were rendered in sRGB color space before being used by each white balance algorithm. Compared to other white balance algorithms \cite{buchsbaum1980spatial, brainard1986analysis, nus}, the proposed CNN achieves desirable results. Fig. \ref{figxyz} shows more qualitative results obtained using the proposed semantic-based white balance.  
\\
In order to investigate the impact of the semantic information on improving the results, we fed the trained model with a correct and a fake semantic mask. As shown in Fig. \ref{fig3}, the incorrect semantic mask leads to undesirable result. Fig. \ref{fig:teaser} shows our results using an sRGB image with color casts. Refinenet \cite{refnet} was used to generate the semantic masks which are not very accurate. Nevertheless, the result is still good, even with an inaccurate semantic mask.

\section{Conclusion}
In this work, we have proposed a semantic-based CNN for color constancy. The proposed CNN uses the input image and a semantic mask in order to estimate the illuminant color and gamma correction parameter. We have shown that, feeding the CNN with semantic information leads to a considerable improvement in the results compared to the results obtained using solely the spatial information of the images. Even in cases where the semantic mask is inaccurate, the provided semantic information can still improve the results of color constancy. 

\bibliographystyle{acm}

\end{document}